\documentclass{article}
\usepackage{spconf,amsmath,graphicx}

\usepackage{epsfig,epsf,fancybox}
\usepackage{amsthm}
\usepackage{amsmath}

\usepackage{tablefootnote}
\usepackage{mathrsfs}
\usepackage{amssymb}
\usepackage{amsfonts}
\usepackage{color}
\usepackage{boxedminipage}
\usepackage{stmaryrd}
\usepackage{multirow}
\usepackage{booktabs}
\usepackage{algorithm}
\usepackage{algorithmic}
\usepackage{cite}
\usepackage{float}
\usepackage{bm}
\usepackage{xcolor}
\usepackage{tcolorbox}
\usepackage{braket,mathdots}
\usepackage{mathtools}
\usepackage[thinlines]{easytable}
\usepackage{array}
\usepackage{bbding}
\usepackage{subcaption}
\usepackage{longtable}

\usepackage{tikz}
\usetikzlibrary{positioning, arrows.meta}
\usetikzlibrary{calc}
\usepackage{tikz}
\usetikzlibrary{shapes.geometric, arrows.meta, positioning, calc, backgrounds}
\usepackage{microtype}
\usepackage{nth}
\usepackage{hyperref}

\title{Distributionally Robust BEST-RQ for Speech Representation Learning}
\name{
\setlength{\tabcolsep}{12pt}
\begin{tabular}{cccc}
Muhammad Khan$^{\star}$ &
Xiaodong Cui$^{\dagger}$ &
Brian Kingsbury$^{\dagger}$ &
George Saon$^{\dagger}$ \\[3pt]
\small \texttt{khanm7@rpi.edu} &
\small \texttt{cuix@us.ibm.com} &
\small \texttt{bedk@us.ibm.com} &
\small \texttt{gsaon@us.ibm.com}
\end{tabular}
\thanks{This work was done while the first author was a visiting student at IBM Research through the RPI-IBM Future of Computing Research Collaboration.}
}

\address{
$^{\star}$Rensselaer Polytechnic Institute \qquad
$^{\dagger}$IBM Research
}

\begin{document}
\ninept
\maketitle
\begin{abstract}
BEST-RQ, based on random projection and quantization, has been a well-established self-supervised learning approach to learn speech representations from large-scale unlabeled audio due to its high performance and efficiency.  However, in BEST-RQ representations learned by random quantization can be sensitive to the choice of quantization targets. To address this issue, we propose a distributionally robust extension of BEST-RQ that seeks representations robust across multiple quantization configurations. Specifically, we define groups by codebooks and formulate the training as a group Distributionally Robust Optimization (DRO) problem. Rather than optimizing solely for the average loss across groups, our approach adaptively reweights groups to emphasize those performing poorly. Additionally, we regularize DRO with a Kullback–Leibler divergence to prevent excessive distribution deviation. We evaluate our approach on the Multilingual LibriSpeech  dataset and show that the distributionally robust BEST-RQ achieves superior performance over the conventional uniform average across multiple codebooks. 

\end{abstract}
\vspace{-0.2cm}
\section{Introduction}
\label{sec:intro}
\vspace{-0.1cm}
In recent years, the limited availability of labeled {speech} data has motivated the development of self-supervised learning (SSL) methods that exploit abundant unlabeled speech. {This gives rise to a commonly-used two-stage strategy where an acoustic model is first pre-trainined using SSL and then fine-tuned using labeled data in downstream tasks.} These SSL methods have progressively explored different pre-training objectives for learning effective speech representations.  wav2vec 2.0 \cite{baevski2020wav2vec} demonstrated that effective speech representations can be learned by masking latent speech representations and predicting quantized targets through contrastive learning. HuBERT \cite{hsu2021hubert} further explored masked prediction by using discrete hidden speech units as learning targets, while W2v-BERT \cite{chung2021w2v} combined contrastive learning with masked language modeling to learn contextualized speech representations. Data2vec \cite{baevski2022data2vec} subsequently introduced contextualized latent targets as a general self-supervised learning objective, while WavLM \cite{chen2022wavlm} jointly learned masked speech prediction and denoising to improve representations for speech content and non-ASR tasks. These developments demonstrate the effectiveness of SSL for learning transferable speech representations from unlabeled audio and motivate further investigation into efficient and robust target construction strategies.
\vspace{-0.1cm}
\subsection{Motivation}
Among the mentioned approaches, BEST-RQ (BERT-based Speech Pre-Training with Random-projection Quantizer) is an effective SSL approach that constructs discrete prediction targets using a fixed random projection and codebook  \cite{chiu2022self}. A key advantage of this approach is that the quantizer does not need to be learned during pre-training, avoiding the additional computational cost associated with learning a codebook. However, fixing a randomly initialized quantizer also introduces a potential source of variability: the resulting prediction task depends on the particular random projection and codebook used to construct the discrete targets.

This dependence is important because the quality of the pre-training task is closely related to how effectively the codebook is utilized.  It has been noted that higher and more uniform codebook utilization is strongly correlated with better pre-training quality, and further highlights that different random initializations of codebook can result in different codebook utilization patterns \cite{chiu2022self}. Consequently, while a fixed random codebook provides a computationally efficient alternative to learned quantization, an individual random configuration need not necessarily be an optimal choice. A representation optimized for one configuration may thus become sensitive to the target space induced by that specific codebook.

A natural alternative is to expose the model to multiple random codebooks during pre-training \cite{zhang2023google}. However, simply training with multiple codebooks using an equal-weight objective implicitly assumes that all codebook-induced prediction tasks are equally informative and equally reliable. This assumption may not hold in practice: different codebooks can induce prediction tasks with different levels of codebook utilization and difficulty, causing some configurations to contribute more to the average training objective than others. Consequently, uniform multi-codebook training does not explicitly encourage robustness to the less favorable configurations and may still favor a subset of codebooks.

These observations motivate the following question: \textbf{Can we learn speech representations that are robust across multiple random codebooks while accounting for the differing quality of the corresponding prediction tasks?} One alternative would be to independently pre-train models with different codebooks and select the best-performing configuration. Although this can identify a favorable target space, it requires repeating the computationally expensive pre-training process and and requires selecting the best codebook. Simply averaging the objectives across multiple codebooks avoids this repeated training cost, but does not explicitly protect against poorly performing or under-utilized configurations.
\vspace{-0.1cm}
\subsection{Contributions}
In this work, we address the aforementioned problem through distributionally robust optimization (DRO). Drawing on Group DRO \cite{sagawa2019distributionally}, we regard each random codebook as a distinct group corresponding to a different discrete prediction task. Rather than weighting all codebooks uniformly, our objective explicitly accounts for their relative training losses and emphasizes configurations on which the representation performs less favorably. We further incorporate KL-divergence regularization to prevent the optimization from becoming excessively concentrated on a single codebook and to maintain a balanced contribution from multiple configurations. This formulation therefore encourages the learned representation to perform consistently across diverse codebook induced target spaces.

The proposed framework retains the computational advantage of fixed random quantization while avoiding reliance on a single randomly selected codebook. More importantly, it goes beyond the {conventional} multi-codebook training by explicitly optimizing for robustness across different prediction tasks. Thus, instead of repeatedly searching for the best codebook or treating all codebooks as equally informative, our approach seeks to learn a representation that remains effective across a range of possible random target spaces.

\section{Background}
\label{sec:format}

\subsection{BEST--RQ}

BEST-RQ constructs discrete prediction targets for speech frames using a fixed random projection matrix and a fixed {random} codebook. Each speech feature is first transformed into a lower-dimensional representation using the random projection matrix and then quantized by selecting the nearest vector in the codebook. The index of the selected codebook entry serves as the discrete target for the corresponding time frame. During pre-training, a subset of speech frames is masked, and the model is trained to predict the corresponding discrete targets for the masked frames.

We now formulate the BEST-RQ objective. Let the $i$-th unlabeled {utterance} be denoted by $\boldsymbol{\xi}^{i}=(x^{i},m^{i})$ where $x^{i}=(x_1^{i},x_2^{i},\ldots,x_{T_i}^{i})$ is a sequence of $T_i$ acoustic feature vectors (obtained after applying mel-spectogram), with $x_t^i\in\mathbb{R}^{d_x}$. Let $m^{i}=(m_1^{i},m_2^{i},\ldots,m_{T_i}^{i}),\  m_t^i\in\{0,1\}$, denotes the masking indicators. A value of $m_t^i=1$ indicates that the
$t$-th frame is masked.

For a given BEST-RQ quantization configuration, the {acoustic} features are
first projected onto a lower dimensional space using a fixed random
projection matrix and subsequently quantized using a fixed codebook.  This
produces a discrete target $y_t^i \in [\mathcal{S}]$, where $S$ is the number of codebook entries,  for each {frame}. We denote the resulting sequence of targets by $y^{i} = \left(y_{1}^{i}, y_{2}^{i}, \ldots, y_{T_{i}}^{i}\right) \in [\mathcal{S}]^{T_{i}}$.  After obtaining the quantization targets, we replace the masked {acoustic} vectors with noise. Specifically, let $\gamma_{t}^{i}$ denote the noise vector used to replace the $i^{th}$ {frame}. The resulting masked input sequence is defined as 
\[
\tilde{x}^{i}_{t} =
\begin{cases}
    \gamma_{t}^{i}, & \text{if } m_{t}^{i}=1,\\
    x_{t}^{i}, & \text{otherwise}.
\end{cases}
\]
The resulting pseudo-labeled example is denoted by $\tilde{\bm{\xi}}^{i} = \left(\tilde{x}^{i}, y^{i}, m^{i}\right)$.
Let
$f_{\theta}(\tilde{x}_{t}^{i})\in\mathbb{R}^{S}
$
denote the model output for frame $t$, 
The BEST-RQ loss for the $i$-th {utterance} is then
\[
\hat{\mathcal{L}}^{B}
(\theta;\tilde{\bm{\xi}}^{i})
=\sum_{t=1}^{T_i}
m_t^i\,\ell\left(f_{\theta}(\tilde{x}_{t}^{i}),y_t^i\right),
\]
where $\ell(\cdot,\cdot)$ denotes the cross entropy loss. Given $N$ unlabeled {utterances}, the empirical BEST-RQ objective is thus, 
\[
\mathcal{L}^{B}(\theta)
=
\frac{1}{\sum_{i=1}^{N}\sum_{t=1}^{T_{i}}m_{t}^{i}}
\sum_{i=1}^{N}
\hat{\mathcal{L}}^{B}
(\theta;\tilde{\bm{\xi}}^{i}).
\]

\subsection{DRO}

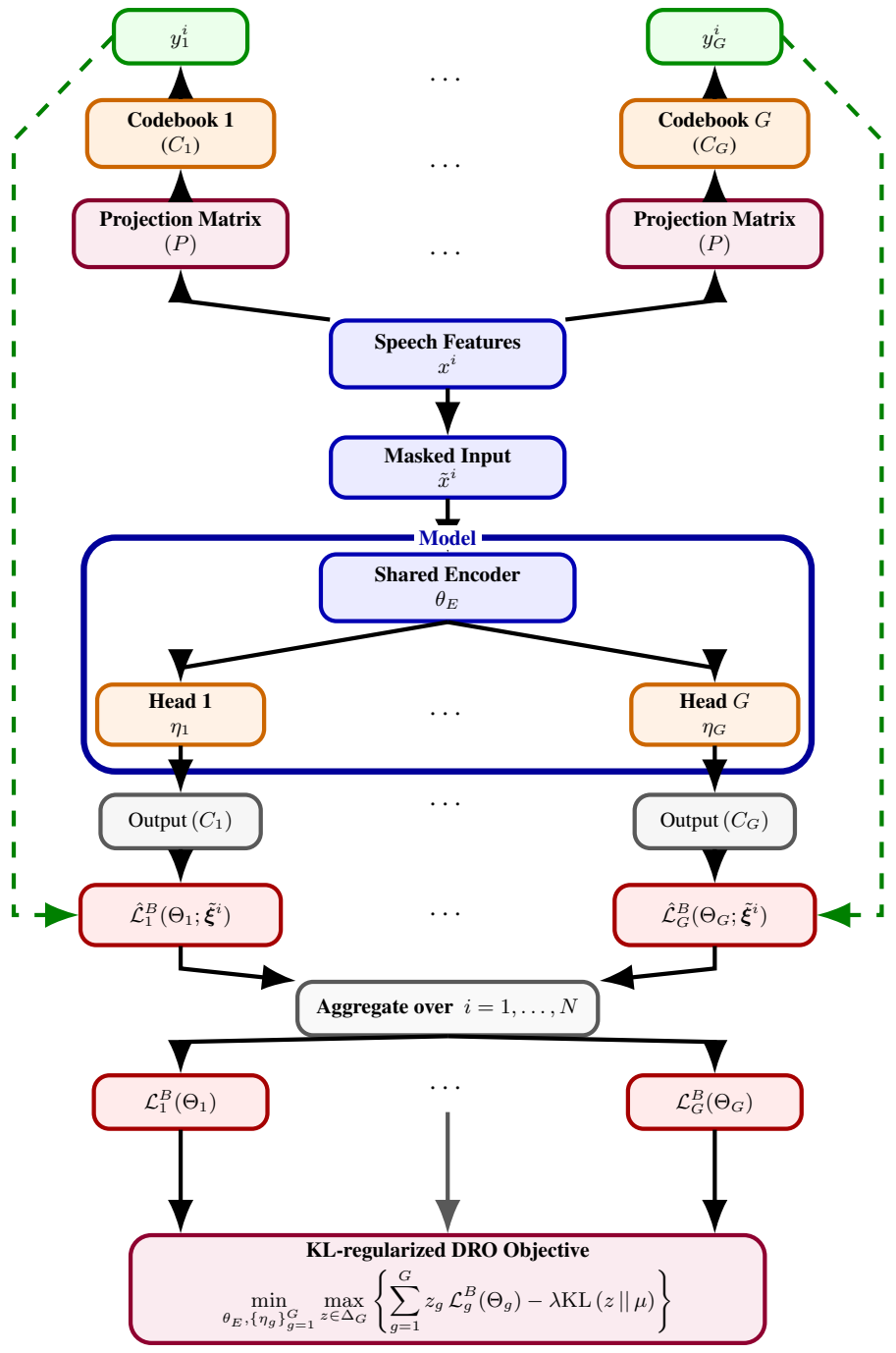
\begin{figure}[t]
\centering

\begin{tikzpicture}[
    >=Latex,
    thick,
    scale=0.45,
    transform shape
]


\node[
    draw=green!55!black,
    fill=green!8,
    rounded corners=3pt,
    minimum width=20mm,
    minimum height=8mm,
    align=center
] (Y1) at (-4,0.25)
{$y_1^i$};

\node[
    font=\large
] at (0,-0.4)
{$\cdots$};

\node[
    draw=green!55!black,
    fill=green!8,
    rounded corners=3pt,
    minimum width=20mm,
    minimum height=8mm,
    align=center
] (YG) at (4,0.25)
{$y_G^i$};


\node[
    draw=orange!80!black,
    fill=orange!12,
    rounded corners=3pt,
    minimum width=28mm,
    minimum height=10mm,
    align=center
] (C1) at (-4,-1.2)
{\textbf{Codebook 1} \\ ($C_1$)};

\node[
    font=\large
] at (0,-1.7)
{$\cdots$};

\node[
    draw=orange!80!black,
    fill=orange!12,
    rounded corners=3pt,
    minimum width=28mm,
    minimum height=10mm,
    align=center
] (CG) at (4,-1.2)
{\textbf{Codebook $G$} \\ ($C_G$)};


\node[
    draw=purple!70!black,
    fill=purple!8,
    rounded corners=3pt,
    minimum width=32mm,
    minimum height=10mm,
    align=center
] (P1) at (-4,-2.7)
{\textbf{Projection Matrix}\\$(P)$};

\node[
    font=\large
] at (0,-3.0)
{$\cdots$};

\node[
    draw=purple!70!black,
    fill=purple!8,
    rounded corners=3pt,
    minimum width=32mm,
    minimum height=10mm,
    align=center
] (PG) at (4,-2.7)
{\textbf{Projection Matrix}\\$(P)$};


\node[
    draw=blue!70!black,
    fill=blue!8,
    rounded corners=3pt,
    minimum width=35mm,
    minimum height=10mm,
    align=center
] (features) at (0,-4.5)
{\textbf{Speech Features}\\$x^i$};


\draw[->]
    (features.north west)
    -- (-4,-3.7)
    -- (P1.south);

\draw[->]
    (features.north east)
    -- (4,-3.7)
    -- (PG.south);


\draw[->]
    (P1.north)
    -- (C1.south);

\draw[->]
    (PG.north)
    -- (CG.south);


\draw[->]
    (C1.north)
    -- (Y1.south);

\draw[->]
    (CG.north)
    -- (YG.south);
    

\node[
    draw=blue!70!black,
    fill=blue!8,
    rounded corners=3pt,
    minimum width=35mm,
    minimum height=9mm,
    align=center
] (masked) at (0,-6.2)
{\textbf{Masked Input}\\$\tilde{x}^i$};

\draw[->]
    (features.south)
    -- (masked.north);


\node[
    draw=blue!70!black,
    fill=blue!8,
    rounded corners=3pt,
    minimum width=38mm,
    minimum height=10mm,
    align=center
] (encoder) at (0,-8.0)
{\textbf{Shared Encoder}\\$\theta_E$};

\draw[->]
    (masked.south)
    -- (encoder.north);


\node[
    draw=orange!80!black,
    fill=orange!10,
    rounded corners=3pt,
    minimum width=25mm,
    minimum height=9mm,
    align=center
] (H1) at (-4,-9.9)
{\textbf{Head 1}\\$\eta_1$};

\node[
    font=\large
] at (0,-9.9)
{$\cdots$};

\node[
    draw=orange!80!black,
    fill=orange!10,
    rounded corners=3pt,
    minimum width=25mm,
    minimum height=9mm,
    align=center
] (HG) at (4,-9.9)
{\textbf{Head $G$}\\$\eta_G$};

\draw[->]
    (encoder.south)
    -- (-4,-9.2)
    -- (H1.north);

\draw[->]
    (encoder.south)
    -- (4,-9.2)
    -- (HG.north);


\begin{scope}[on background layer]

\node[
    draw=blue!60!black,
    very thick,
    rounded corners=6pt,
    minimum width=10.9cm,
    minimum height=3.5cm
] at (0,-9.0) {};

\end{scope}

\node[
    font=\bfseries\small,
    text=blue!65!black,
    fill=white,
    inner sep=2pt
] at (0,-7.25)
{Model};


\node[
    draw=gray!70!black,
    fill=gray!7,
    rounded corners=3pt,
    minimum width=24mm,
    minimum height=8mm
] (O1) at (-4,-11.5)
{$\text{Output} \, (C_1)$};

\node[
    font=\large
] at (0,-11.25)
{$\cdots$};

\node[
    draw=gray!70!black,
    fill=gray!7,
    rounded corners=3pt,
    minimum width=24mm,
    minimum height=8mm
] (OG) at (4,-11.5)
{$\text{Output} \, (C_G)$};

\draw[->]
    (H1.south)
    -- (O1.north);

\draw[->]
    (HG.south)
    -- (OG.north);


\node[
    draw=red!65!black,
    fill=red!8,
    rounded corners=3pt,
    minimum width=30mm,
    minimum height=9mm,
    align=center
] (l1) at (-4,-12.9)
{$\hat{\mathcal{L}}_1^B(\Theta_1;\tilde{\boldsymbol{\xi}}^i)$};

\node[
    font=\large
] at (0,-12.9)
{$\cdots$};

\node[
    draw=red!65!black,
    fill=red!8,
    rounded corners=3pt,
    minimum width=30mm,
    minimum height=9mm,
    align=center
] (lG) at (4,-12.9)
{$\hat{\mathcal{L}}_G^B(\Theta_G;\tilde{\boldsymbol{\xi}}^i)$};

\draw[->]
    (O1.south)
    -- (l1.north);

\draw[->]
    (OG.south)
    -- (lG.north);


\draw[
    ->,
    dashed,
    green!50!black
]
    (Y1.west)
    -- (-6.5,-1.3)
    -- (-6.5,-12.9)
    -- (l1.west);

\draw[
    ->,
    dashed,
    green!50!black
]
    (YG.east)
    -- (6.5,-1.3)
    -- (6.5,-12.9)
    -- (lG.east);


\node[
    draw=gray!70!black,
    fill=gray!5,
    rounded corners=3pt,
    minimum width=45mm,
    minimum height=8mm,
    align=center
] (aggregate) at (0,-14.3)
{\textbf{Aggregate over } $i=1,\ldots,N$};

\draw[->]
    (l1.south)
    -- (-4,-13.7)
    -- (aggregate.north west);

\draw[->]
    (lG.south)
    -- (4,-13.7)
    -- (aggregate.north east);


\node[
    draw=red!65!black,
    fill=red!8,
    rounded corners=3pt,
    minimum width=26mm,
    minimum height=8mm,
    align=center
] (L1) at (-4,-15.7)
{$\mathcal{L}_1^B(\Theta_1)$};

\node[
    font=\large
] at (0,-15.5)
{$\cdots$};

\node[
    draw=red!65!black,
    fill=red!8,
    rounded corners=3pt,
    minimum width=26mm,
    minimum height=8mm,
    align=center
] (LG) at (4,-15.7)
{$\mathcal{L}_G^B(\Theta_G)$};

\draw[->]
    (aggregate.south)
    -- (-4,-14.8)
    -- (L1.north);

\draw[->]
    (aggregate.south)
    -- (4,-14.8)
    -- (LG.north);


\node[
    draw=purple!75!black,
    fill=purple!8,
    rounded corners=4pt,
    minimum width=95mm,
    minimum height=14mm,
    align=center
] (DRO) at (0,-18.5)
{
    \textbf{KL-regularized DRO Objective}\\[2pt]
    $\displaystyle
    \min_{\theta_E,\{\eta_g\}_{g=1}^{G}}
    \max_{z\in\Delta_G}
    \left\{
    \sum_{g=1}^{G} z_g\,\mathcal{L}_g^B(\Theta_g)
    - \lambda \mathrm{KL}\left(z \,||\, \mu\right)
    \right\}
    $
};

\draw[->]
    (L1.south)
    -- (L1.south |- DRO.north);

\draw[->]
    (LG.south)
    -- (LG.south |- DRO.north);
    
\draw[
    ->,
    gray!70!black
]
    (0,-15.85)
    -- (0,-16.4)
    -- (DRO.north);

\end{tikzpicture}

\caption{
{Overview of the proposed DRO-BEST-RQ with KL regularization. For an utterance $i$, the acoustic features $x^i$ are first projected into a lower-dimensional space using a shared projection matrix $P$. The resulting representations are then quantized using multiple fixed random codebooks $C_g$ to generate group-specific pseudo-labels $y_g^i$.  The input acoustic features are processed by a shared encoder and group-specific prediction heads.  The cross-entropy loss is computed for each codebook on those masked frames which are then combined through the KL-regularized DRO objective.}
}\vspace{-0.3cm}


\label{fig:multi_codebook_DRO}

\end{figure}

DRO considers optimization objectives that account for variation across a collection of conditions rather than focusing solely on average performance. Group DRO extends this notion by considering multiple groups and adaptively weighting them according to their associated losses, placing greater emphasis on groups for which the model performs poorly \cite{sagawa2019distributionally}.

We now formulate the DRO objective. Let $\mathcal{G}=\{1,\ldots,G\}$ denote a collection of groups, and let $\mathcal{L}_g(\theta)$ denote the loss associated with group $g$. The Group DRO objective can be expressed as
\[
\min_{\theta}
\max_{z\in\Delta_G}
\sum_{g=1}^{G}
z_g\mathcal{L}_g(\theta),
\]
where $z=\left(z_1,\ldots,z_{G}\right)$ denotes a probability distribution over the groups and
\[
\Delta_{G}=
\left\{
z\in\mathbb{R}^{G}:
z_g\geq0,\;
\sum_{g=1}^{G}z_g=1
\right\}
\]
is the probability simplex. The distribution $z$ determines the relative contribution of each group to the objective.

{To regularize the group distribution, we introduce KL-divergence regularization with respect to a uniform reference distribution. 
\[
\min_{\theta}
\max_{z\in\Delta_G}
\left[
\sum_{g=1}^{G}
z_g\mathcal{L}_g(\theta)
-
\lambda D_{\mathrm{KL}}(z\|u)
\right],
\]
where $\mu$ denote the uniform distribution over the $G$ groups; $\lambda>0$ is the regularization parameter. The KL term penalizes deviations of the group distribution from the uniform reference distribution, thereby allowing us to control the extent of group reweighting.}

\section{PROBLEM FORMULATION}

Having introduced BEST-RQ and DRO, we now formulate our learning problem. In our setting, each group $g$ is associated with a distinct codebook, which determines its BEST-RQ quantization configuration and corresponding discrete prediction targets. We would like to emphasize that the groups do not correspond to different subsets of the speech data. Instead, the same {utterances} are processed under each quantization configuration, resulting in different target spaces for the same input speech.

The framework is illustrated in Figure \ref{fig:multi_codebook_DRO}. The model consists of a shared encoder backbone and a group-specific prediction heads. Let $\theta_E$ denote the parameters of the shared encoder, and let $\eta_g$ denote the parameters of the prediction head associated with group $g$. We define the parameters involved in group $g$ as
\[
\Theta_g = \left(\theta_E,\eta_g\right),\qquad g\in\mathcal{G}.
\]

Thus, the encoder is jointly optimized across all BEST--RQ configurations, while each prediction head specializes to the target space induced by its corresponding random codebook.

Let $\mathcal{L}^{B}_g(\Theta_g)$ denote the BEST--RQ loss for group $g$. We formulate the learning problem as follows:
\begin{align}
\min_{\theta_E,\{\eta_g\}_{g=1}^{G}}
\max_{z\in\Delta_G}
\left\{\sum_{g=1}^{G}
z_g\mathcal{L}^{B}_g(\Theta_g)
-\lambda D_{\mathrm{KL}}(z\|\mu).\right\}
\label{eq:proposed_objective}
\end{align}
\vspace{-0.4cm}


\section{Algorithm}

The pseudocode for DRO-BEST-RQ is provided in Algorithm \ref{alg:bestrq_ctc_maml_multi_codebook}. Since the problem is unconstrained with respect to the model parameters
$\theta_E$ and $\{\eta_g\}_{g=1}^{G}$, while the adversarial variable
$z$ is constrained to the probability simplex $\Delta_G$, we solve the
 minimax problem using a minibatch stochastic gradient
descent--projected stochastic gradient ascent (SGD-PSGA) procedure.
At each iteration, we estimate the gradients of both the minimization
and maximization objectives using the same minibatch, update the model
parameters via stochastic gradient descent, and update $z$ via
stochastic gradient ascent followed by projection onto $\Delta_G$. Specifically, for an arbitrary point $w \in \mathbb{R}^{G}$, the projection onto
$\Delta_G$ is defined as
\begin{equation*}
\text{Proj}_{\Delta_G}(w)
=
\underset{q \in \Delta_G}{\arg\min}\;
\frac{1}{2}\left\|w-q\right\|_2^2.
\label{eq:simplex_projection}
\end{equation*}
\vspace{-0.3cm}

\begin{algorithm}[h]
\caption{DRO-BEST-RQ}
\label{alg:bestrq_ctc_maml_multi_codebook}
\begin{algorithmic}[1]
    \STATE \textbf{Inputs:} $\alpha, \beta, K > 0$, $\theta_{E}^{0}$, $\left\{\eta_{g}^{0}\right\}_{g=1}^{n}$, $z^{0}$ 
\FOR{$k = 0, 1, 2, \ldots, K-1$}

\STATE \textbf{Data:}
\STATE Sample a minibatch $A^{k} = \left\{\bm{\xi}^{0}, \ldots, \bm{\xi}^{P-1}\right\} $

\FOR{$g = 0, 1, 2, \ldots, G - 1 $} 
\STATE Apply the label generation procedure on each sample in the minibatch to yield
\STATE $\tilde{A}^{k}_{g} \gets \left\{\tilde{\bm{\xi}}^{0}_{g}, \ldots, \tilde{\bm{\xi}}^{P-1}_{g}\right\} $
\ENDFOR

\STATE \textbf{Gradient computation:}
\FOR{$g = 0, 1, 2, \ldots, G - 1 $} 
\STATE $Q_{g}\left(\theta_{E}^{k}; \tilde{A}^{k}_{g}\right) \gets \frac{\sum_{i=1}^{P} \nabla_{\theta_{E}}\hat{\mathcal{L}}_{g}^{B}\left(\theta_{E}^{k}, \eta_{g}^{k}; \tilde{\bm{\xi}}^{i}_{g}\right)}{\sum_{i=0}^{P-1}\sum_{t=0}^{T_{i} - 1}m_{t}^{i}}$
\STATE $R_{g}\left(\eta_{g}^{k}\,; \tilde{A}^{k}_{g}\right) \gets \frac{\sum_{i=1}^{P} \nabla_{\eta_{g}}\hat{\mathcal{L}}_{g}^{B}\left(\theta_{E}^{k}, \eta_{g}^{k}; \tilde{\bm{\xi}}^{i}_{g}\right)}{\sum_{i=0}^{P-1}\sum_{t=0}^{T_{i} - 1}m_{t}^{i}}$
\ENDFOR

\FOR{$j = 0, 1, 2, \ldots, G-1$}
\STATE $H_{j}^{k} \gets \frac{\sum_{i=1}^{P} \hat{\mathcal{L}}_{j}^{B}\left(\theta_{E}^{k}, \eta_{j}^{k}; \tilde{\bm{\xi}}^{i}_{j}\right)}{\sum_{i=0}^{P-1}\sum_{t=0}^{T_{i} - 1}m_{t}^{i}}$
\STATE $\left[U\left(z^{k}; \tilde{A}^{k}\right)\right]_{j} \gets H_{j}^{k} - \lambda \left(1 + \log\left(\frac{z_{j}}{\mu_{j}}\right)\right)$
\ENDFOR

\STATE \textbf{Encoder parameters update:}

\STATE \[\theta_{E}^{k+1} \gets  \theta_{E}^{k} - \alpha \sum_{g=0}^{G-1}z_{t}^{k}Q_{g}\left(\theta_{E}^{k}; \tilde{A}^{k}\right) \]

\STATE \textbf{Update parameters of codebook specific heads}

\FOR{$g = 0, 1, 2, \ldots, G - 1 $} 
\STATE \[\eta_{g}^{k+1} \gets  \eta_{g}^{k} - \alpha z_{g}^{k}R_{g}\left(\eta_{g}; \tilde{A}^{k}\right) \]
\ENDFOR

\STATE \textbf{Update $z$}

\STATE \[z^{k+1} \gets  \text{Proj}_{\Delta_{n}}\left(z^{k} + \beta U\left(z^{k}; \tilde{A}^{k}\right)\right) \]

\ENDFOR

\STATE \textbf{Output:} $\theta_{E}^{K}$

\end{algorithmic}
\end{algorithm}
\vspace{-0.2cm}

\section{Experimental Results}

\subsection{Dataset}


{We evaluate the proposed DRO-BEST-RQ on the Multilingual Librispeech (MLS) dataset~\cite{Pratap2020}. Five languages are used for the experiments which are English (en), Spanish (es), French (fr), German (de) and Portuguese (pt). There are 44,659.74 hours of English data from which we randomly select 4,000 hours. For the other four languages, we use all the training data in the dataset. The multilingual data from those five languages are pooled together for the BEST-RQ unsupervised training by removing the labels, which amounts to a total of  8,121.73 hours.  The pre-trained BEST-RQ model is then fine-tuned individually for each language under CTC using its own data with labels \cite{graves2006connectionist}. The fine-tuned models are evaluated using test data from each language respectively. Table~\ref{tab:dataset_stats} summarizes the amount of speech data used for pre-training, fine-tuning, and test in each language. Note that for each language the pre-training and fine-tuning data is the same. The difference is whether the labels are used.} \vspace{-0.3cm}

\begin{table}[t]
    \centering
    \scriptsize
    \renewcommand{\arraystretch}{1.2}
    \caption{Statistics of the MLS data used in our experiments.}\vspace{-0.1cm}
    \label{tab:dataset_stats}
    \begin{tabular}{lccc}
        \toprule
        \textbf{Language} & \textbf{Pre-Training (hrs)} & \textbf{Fine-Tuning (hrs)} & \textbf{Test (hrs)} \\
        \midrule
        English    & 4000.00 & 4000.00 & 15.55 \\
        Spanish    & 917.68  & 917.68  & 10.00 \\
        French     & 1076.58 & 1076.58 & 10.07 \\
        German     & 1966.51 & 1966.51 & 14.29 \\
        Portuguese & 160.96  & 160.96  & 3.74 \\
        \bottomrule
    \end{tabular}
    \vspace{-0.3cm}
\end{table}

\subsection{Model Architecture}


{The ASR models we train use the same architecture as \href{https://huggingface.co/ibm-granite/granite-speech-5.0-470m-turboctc}{ibm-granite/granite-speech-5.0-470m-turboctc}. 
In the BEST-RQ training, the model comprises a stack of 16 Conformer blocks taking a 320-dimensional input feature (2 stacked frames of 80-dimensional log Mel spectrogram features with temporal deltas). The Conformer blocks have an internal dimension of 1024, a temporal kernel width of $k=7$, use 8 128-dimensional attention heads, and chunkwise attention~\cite{GoogleUSM} with a block size of 128. The first two Conformer blocks perform a $2\times$ temporal downsampling by applying strided convolutions in the Conformer's depthwise convolutional block and averaging successive pairs of sequence positions in the corresponding residual connection. Thus, the model output rate is 12.5 tokens/sec. The masking is applied by randomly selecting 2\% of the total frame length and masking spans 20 frames. The masked frames are replaced with Gaussian noise of zero mean and 0.1 variance. Each codebook has 1024 entries. When fine-tuning under CTC, the top linear layer is replaced with a randomly initialized new one which is connected to an alphabet of 16,384 BPE tokens, including the obligatory CTC BLANK symbol. In addition, self-conditioning~\cite{Nozaki2021} is added at the \nth{8} layer in the CTC fine-tuning stage.}\vspace{-0.3cm}

\subsection{Implementation Details}

{We first establish supervised baselines for the five languages using the labeled fine-tuning data without pre-training. Then we create two BEST-RQ baselines with a single codebook and four codebooks, respectively. This involves a two-stage process. In the first stage, the acoustic model is pre-trained under BEST-RQ using pooled unlabeled multilingual data. In the second stage, the pre-trained acoustic model is fine-tuned  using labeled fine-tuning data for each language. The multi-codebook BEST-RQ baseline employs the conventional uniformly weighted losses (i.e. assigning a weight of $0.25$ to the loss corresponding to each codebook.). We then investigate the proposed DRO-BEST-RQ with multi-codebooks as illustrated in Fig.\ref{fig:multi_codebook_DRO} and Alg.\ref{alg:bestrq_ctc_maml_multi_codebook} with various $\lambda$'s.}

Both BEST-RQ training and fine-turing are run for 30 epochs with a OneCycleLR scheduler using 8 NVIDIA H100 GPUs. 
{For $\alpha$, we set $\alpha_{\text{max}}=5e^{-4}$, $\alpha_{{\text{init}}}=5e^{-5}$, and $\alpha_{\text{final}}=5e^{-6}$. For $\beta$, we use the same schedule with $\beta_{\text{max}}=1e^{-3}$, $\beta_{\text{init}}=1e^{-4}$, and $\beta_{\text{final}}=1e^{-5}$. The warmup is set to $10\%$ of the total training for both $\alpha$ and $\beta$.}
All other architectural and optimization settings are kept fixed across configurations to ensure a fair comparison. A greedy decoder is used 
for inference.

\vspace{-0.2cm}




\subsection{Illustration of the Group DRO Mechanism in practice}

{

To provide an illustration of \eqref{eq:proposed_objective} in practice, Table~\ref{tab:group_dro_mechanism} presents the group-specific BEST-RQ losses and the corresponding DRO weights at an arbitrarily selected training iteration. The values are reported separately for models trained with and without KL regularization and are intended to highlight the behavior of the optimization rather than to represent aggregate or final model performance. For the KL-regularized model $(\lambda > 0)$, the group weights adapt to the relative difficulty of the BEST-RQ configurations: the group with the largest loss receives the largest weight, while the remaining codebooks retain non-zero weights. Consequently, all codebooks continue to contribute to the optimization of the shared encoder, with greater emphasis placed on the more difficult configurations. In contrast, when KL regularization is removed $(\lambda = 0)$, the group distribution concentrates entirely on the highest-loss group in the corresponding independently trained model, which amounts to a single-codebook training scenario but that single codebook is the hardest one in all codebook candidates. For random codebooks, a hard codebook typically represents a broader coverage of the random codewords which could benefit the unsupervised learning. This comparison demonstrates the role of the KL term in controlling the degree of group reweighting. Specifically, it preserves the adaptive behavior of Group DRO while discouraging excessive concentration on a single BEST-RQ configuration.

\begin{table}[t]
\centering
\caption{Illustration of the Group DRO weighting mechanism across four BEST-RQ codebooks at an arbitrarily selected iteration.}
\label{tab:group_dro_mechanism}
\resizebox{\columnwidth}{!}{%
\begin{tabular}{c | c | c | c}
\toprule
Training setting & Group 
& BEST-RQ loss $\left(\mathcal{L}_g\right)$ 
& DRO weight $\left(z_g\right)$ \\
\midrule
\multirow{4}{*}{$\lambda>0$}
& $g = 1$ & 3.4055 & 0.2338 \\
& $g = 2$ & 3.4789 & 0.2517 \\
& $g = 3$ & 3.4152 & 0.2361 \\
& $g = 4$ & \textbf{3.5800} & \textbf{0.2784} \\
\midrule
\multirow{4}{*}{$\lambda=0$}
& $g = 1$ & 3.3118 & 0.0 \\
& $g = 2$ & 3.3089 & 0.0 \\
& $g = 3$ & 3.2886 & 0.0 \\
& $g = 4$ & \textbf{3.3513} & \textbf{1.0} \\
\bottomrule
\end{tabular}%
}
\end{table}
}

\subsection{Results \& Discussion}

We report Word Error Rate (WER) on the  test sets described in Table~\ref{tab:results}. Results are reported for each of the five languages. The WERs of Portuguese are obviously higher than the rest four languages due to its limited fine-tuning data (only 161 hours).  We compare the supervised baseline, the multi-codebook approach with uniformly weighted losses, and our proposed DRO approach with $\lambda \in \left\{0.0, 0.5, 1.0, 1.5\right\}$. {We note that setting $\lambda = 0$ reduces \eqref{eq:proposed_objective} to a worst-case optimization over the codebooks, where, at each iteration, $z$ places all its mass on the codebook with the highest loss, thereby optimizing the model parameters against the hardest codebook.}

\begin{table}[t]
    \centering
    \scriptsize
    \renewcommand{\arraystretch}{1.2}
    \caption{WER (\%) on the  MLS test sets}
    \label{tab:results}
    \begin{tabular}{lcccccc}
        \toprule
        \textbf{Model} & \textbf{en} & \textbf{es} & \textbf{fr} & \textbf{gr} & \textbf{pt} & \textbf{Avg}\\
        \midrule
        Supervised Baseline & 8.0 & 5.1 & 6.8 & 6.3 & 26.0 & 10.4 \\
        BEST--RQ (Single-Codebook)& 8.0 & 4.8 & 6.4 & 6.3 & 20.8 & 9.3  \\
        BEST--RQ (Multi-Codebook) & 7.9 & 4.6 & 6.6 & 6.3 & 18.9 & 8.9\\
        \textbf{DRO--BEST--RQ} ($\lambda=0.0$)   & 8.0 & 4.8 & 6.7 & 6.2 & 20.0 & 9.1\\
        \textbf{DRO--BEST--RQ} ($\lambda=0.5$) & \textbf{7.9} & \textbf{4.4} & \textbf{6.3} & \textbf{6.0} & \textbf{18.5} &\textbf{8.6} \\
        \textbf{DRO--BEST--RQ} ($\lambda=1.0$)   & \textbf{7.7} & \textbf{4.3} & \textbf{6.3} & \textbf{5.9} & \textbf{17.7} & \textbf{8.4}\\
        \textbf{DRO--BEST--RQ} ($\lambda=1.5$) & \textbf{7.9} & \textbf{4.3} & \textbf{6.3} & \textbf{6.1} & \textbf{18.4} & \textbf{8.6} \\
        \bottomrule
    \end{tabular}
    \vspace{-0.3cm}
\end{table}


The results in Table~\ref{tab:results} demonstrates the effectiveness of our proposed Algorithm~\ref{alg:bestrq_ctc_maml_multi_codebook}. While multi-codebook BEST-RQ already improves upon both the supervised baseline and the single-codebook formulation, our algorithm consistently delivers additional gains on top of it for $\lambda \in \left\{0.5, 1.0, 1.5\right\}$. Importantly, these improvements are observed across multiple values of $\lambda$, demonstrating that the gains are not tied to a single carefully selected setting. This highlights the advantage of our DRO formulation, which adaptively reweighs the codebook-specific objectives to reduce the dominance of favorable codebooks and encourage the shared encoder to remain robust to less favorable or under utilized codebook configurations. The strongest overall performance is achieved at $\lambda = 1.0$, while the gains for $\lambda \in \left\{0.5, 1.5\right\}$ further illustrate the effectiveness of our algorithm. In all the experiments, the most significant language-specific gains come from Portuguese due to its limited data, which also indicates that low-resource languages benefit from the SSL pre-training more.  

\section{Concluding Remarks}
In this work, we proposed a KL-regularized distributionally robust optimization (DRO) framework for multi-codebook BEST--RQ. Our framework treats each random codebook as a separate prediction task and adaptively emphasizes less favorable codebook configurations while preventing excessive reweighting through KL regularization. Experiments on MLS data show that our approach consistently outperforms uniformly weighted multi-codebook BEST--RQ across multiple languages, establishing DRO as a principled and effective way to improve BEST--RQ pre-training without relying on a single random codebook assuming that all codebook configurations are equally informative.

\bibliographystyle{IEEEbib}
\bibliography{strings}

\end{document}